\documentclass[final]{cvpr}

\RequirePackage{algorithm}
\RequirePackage{algorithmic}

\usepackage{hyperref}
\hypersetup{
    colorlinks=true,
    linkcolor=green,
    filecolor=magenta,      
    urlcolor=cyan,
}
\usepackage{url}

\usepackage{amsmath}
\usepackage{amssymb}
\usepackage{microtype}
\usepackage{graphicx}
\usepackage{subfigure}
\usepackage{amsfonts}
\usepackage{booktabs}
\usepackage{multirow}
\usepackage{eqparbox}
\usepackage{times}
\usepackage{epsfig}
\usepackage{graphicx}
\usepackage{subfigure}
\usepackage{amsfonts}
\usepackage{booktabs}
\usepackage{multirow}
\usepackage{eqparbox}
\usepackage{epstopdf}
\usepackage[utf8]{inputenc}

\newcommand{\cut}[1]{}

\DeclareMathOperator*{\argmin}{arg\,min}
\newcommand{\bfs}[1]{\textbf{#1}}
\renewcommand\vec{\mathbf}

\usepackage{color, colortbl}
\definecolor{Gray}{gray}{0.85}
\newcolumntype{a}{>{\columncolor{Gray}}c}

\begin{document}

\title{Searching for Robustness: Loss Learning for Noisy Classification Tasks}

\author{Boyan~Gao\textsuperscript{\rm 1}\qquad~Henry~Gouk\textsuperscript{\rm 1}\qquad~Timothy~M.~Hospedales\textsuperscript{\rm 1, 2}\\
\textsuperscript{\rm 1}School of Informatics, University of Edinburgh, United Kingdom\\
\textsuperscript{\rm 2}Samsung AI Centre, Cambridge, United Kingdom\\
\small{\texttt{\{boyan.gao,henry.gouk,t.hospedales\}@ed.ac.uk}}
}

\maketitle

\begin{abstract}
We present a ``learning to learn'' approach for automatically constructing white-box classification loss functions that are robust to label noise in the training data. We parameterize a flexible family of loss functions using Taylor polynomials, and apply evolutionary strategies to search for noise-robust losses in this space. To learn re-usable loss functions that can apply to new tasks, our fitness function scores their performance in aggregate across a range of training dataset and architecture combinations. The resulting white-box loss provides a simple and fast ``plug-and-play'' module that enables effective noise-robust learning in diverse downstream tasks, without requiring a special training procedure or network architecture. The efficacy of our method is demonstrated on a variety of datasets with both synthetic and real label noise, where we compare favorably to previous work.
\end{abstract}
\section{Introduction}
The success of modern deep learning is largely predicated on large amounts of accurately labelled training data. However, training with large quantities of gold-standard labelled data is often not achievable. This is often because professional annotation is too costly to achieve at scale and machine learning practitioners resort to less reliable crowd-sourcing, web-crawled incidental annotations~\cite{chen2015weblyCNN}, or imperfect machine annotation~\cite{kuznetsova2020openImages}; while in other situations the data is hard to classify reliably even by human experts, and label-noise is inevitable. These considerations have led to a large and rapidly progressing body of work focusing on developing noise-robust learning approaches~\cite{ren2018learning,han2018co}. Diverse solutions have been studied including those that modify the training algorithm through teacher-student~\cite{jiang2018mentornet,han2018co} learning, or identify and down-weight noisy instances~\cite{ren2018learning}. Much simpler, and therefore more widely applicable, are attempts to define noise-robust loss functions that provide drop-in replacements for standard losses such as cross-entropy~\cite{wang2019symmetric,zhang2018generalized,ghosh2017robust}. These studies hand engineer robust losses, motivated by different considerations including risk minimisation~\cite{ghosh2017robust} and information theory~\cite{xu2019l_dmi}. In this paper we explore an alternative data-driven approach~\cite{hutter2019autoMLbook} to loss design, and search for a simple white-box function that provides a general-purpose noise-robust drop-in loss.

\begin{figure}
    \centering
    \includegraphics[width=1.0\columnwidth]{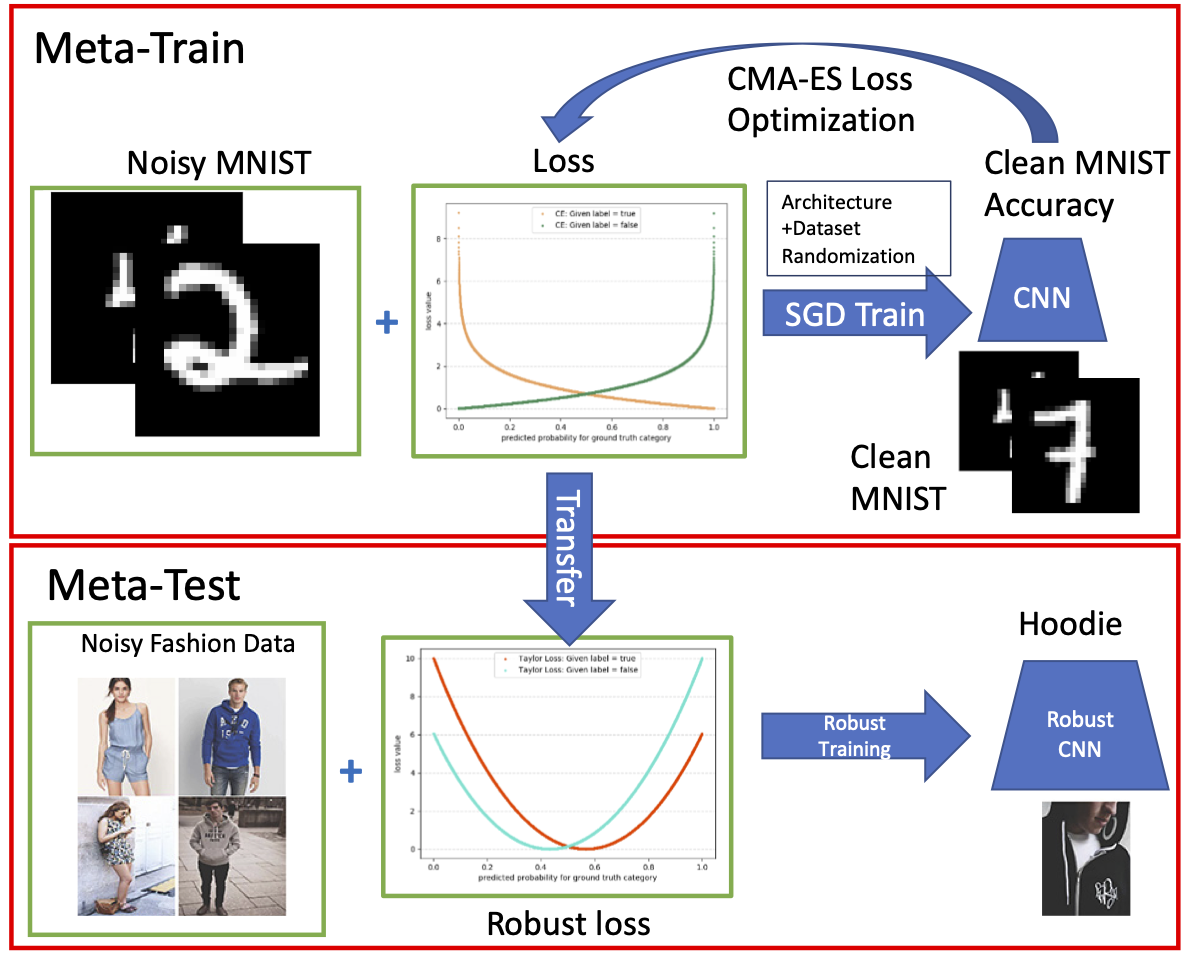}
    \caption{Schematic of our robust loss search framework. (1) We train a robust loss function so as to optimize validation performance of a CNN trained with synthetic label noise using this loss. (2) Thanks to dataset and architecture randomization, our learned loss is reusable and can be deployed to new tasks, including those without clean validation set to drive robust learning.}
    \label{fig:teaser}
\end{figure}
We perform evolutionary search on a space of loss functions parameterised as Taylor polynomials. Every function in this space is smooth and differentiable, and thus provides a valid loss that can be easily plugged into existing deep learning frameworks. Meanwhile, this search space provides a good trade-off between the flexibility to represent non-trivial losses, and a  low-dimensional white-box parameterisation that is efficient to search and reusable across tasks without overfitting. To score a given loss during our search, we use it to train neural networks on noisy data, and then evaluate the clean validation performance of the trained model. To learn a general purpose loss, rather than one that is specific to a given architecture or dataset, we explore domain randomisation~\cite{tobin2017domain} in the space of architectures and datasets. Scoring losses according to their average validation performance in diverse conditions leads to reusable functions that can be applied to new datasets and architectures, as illustrated in Figure~\ref{fig:teaser}.

We apply our learned loss function to train various MLP and CNN architectures on several benchmarks including MNIST, FashionMNIST, USPS, CIFAR-10, and CIFAR-100 with different types of simulated label noise. We also test our loss on a large real-world noisy  label dataset, Clothing1M. The results verify the re-usability of our learned loss and its efficacy compared to state-of-the-art in a variety of settings. An important advantage of our approach compared to previous work that makes use of AutoML techniques to learn noise-robust loss functions is transferability. Previous methods for tackling noisy classification tasks often require a clean (i.e., noiseless) validation dataset to use as a meta-supervision signal~\cite{ren2018learning,shu2019metaWeightNet}. In contrast, we formulate our learning algorithm in such a way that we can instead make use of a clean validation set on an arbitrary auxiliary domain that can be completely different from the domain of interest.

\begin{figure*}[!t]
\begin{minipage}[b]{.32\linewidth}
 \centering
 \centerline{\epsfig{figure ={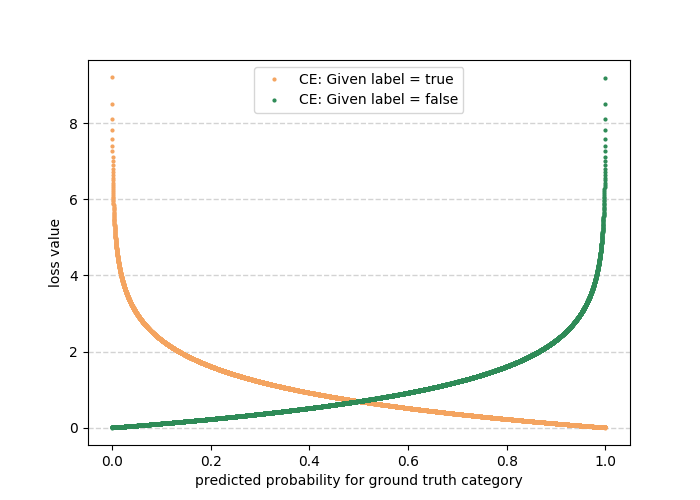}, width= 5cm}}
\end{minipage}
\hfill
\begin{minipage}[b]{.32\linewidth}
  \centering
 \centerline{\epsfig{figure={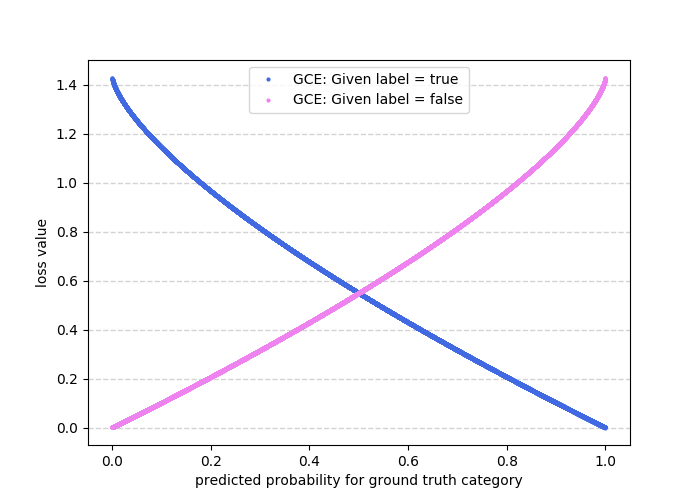},width=5cm}}
\end{minipage}
\hfill
\begin{minipage}[b]{.32\linewidth}
  \centering
 \centerline{\epsfig{figure={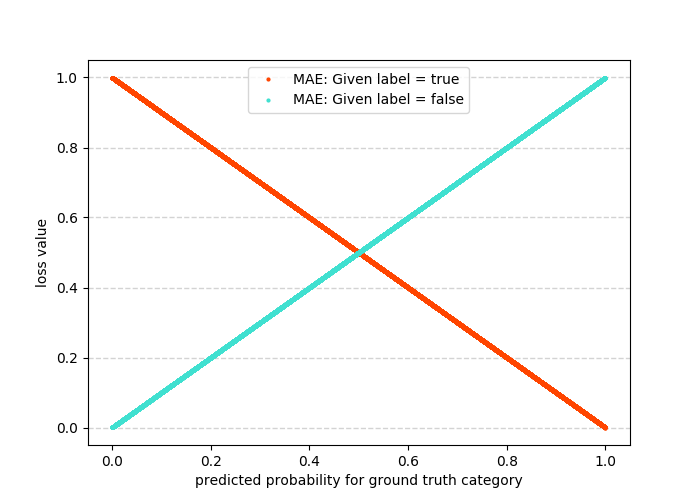},width=5cm}}
\end{minipage}
\vfill
\begin{minipage}[b]{.32\linewidth}
  \centering
 \centerline{\epsfig{figure={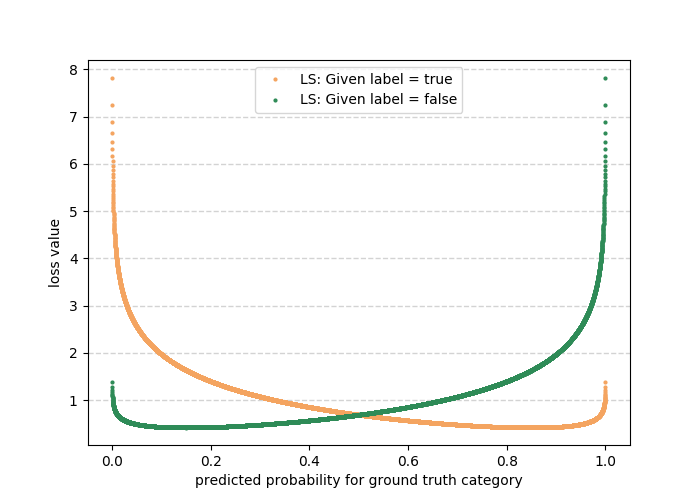},width=5cm}}
\end{minipage}
\hfill
\begin{minipage}[b]{.32\linewidth}
  \centering
 \centerline{\epsfig{figure={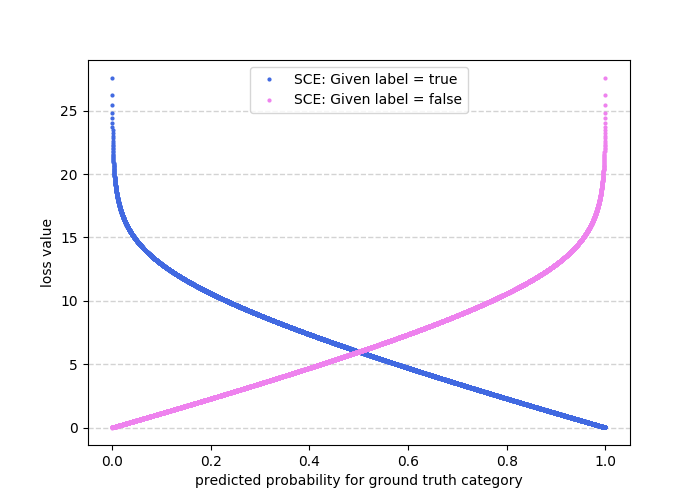},width=5cm}}
\end{minipage}
\hfill
\begin{minipage}[b]{.32\linewidth}
  \centering
 \centerline{\epsfig{figure={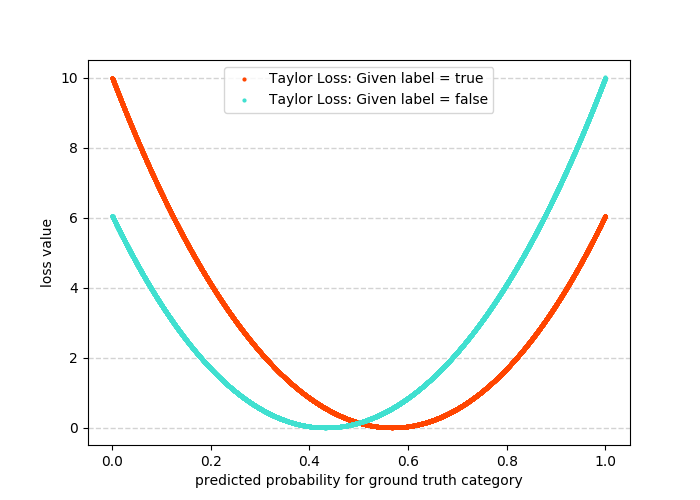},width=5cm}}
\end{minipage}
\caption{Existing hand-designed robust losses and our meta-learned robust loss. Top left: Conventional Cross-Entropy (CE); Top middle: Generalised Cross Entropy (GCE)~\cite{zhang2018generalized}; Top right: Mean Absolute Error (MAE)~\cite{ghosh2017robust}; Bottom left: label-smoothing~\cite{pereyra2017regularizing}. Bottom middle: Symmetric Cross Entropy~\cite{wang2019symmetric}. Bottom right: Our learned loss.}
\label{learned_loss_fuc_binary}
\end{figure*}

\section{Related Work}
\textbf{Label Noise}\quad Learning with label noise is now a large research area due to its practical importance. Song \etal~\cite{song2020labelNoiseSurvey} present a detailed survey explaining the variety of approaches previously studied including designing noise robust neural network architectures \cite{chen2015weblyCNN}, regularisers such as label-smoothing~\cite{szegedy2016rethinking,pereyra2017regularizing}, sample selection methods that attempt to filter out noisy samples -- often by co-teaching or student teacher learning with multiple neural networks ~\cite{jiang2018mentornet,han2018co,wei2020combating}, various meta-learning approaches that often aim to down-weight noisy samples using meta-gradients from a validation set~\cite{ren2018learning,shu2019metaWeightNet}, and robust loss design. Among these families of approaches, we are motivated to focus on robust loss design due to simplicity and general applicability -- we wish to use standard architectures and standard learning algorithms. Major robust losses include MAE, shown to be theoretically robust in \cite{ghosh2017robust}, but hard to train in~\cite{zhang2018generalized}; GCE which attempts to be robust yet easy to train \cite{zhang2018generalized}, and symmetric cross-entropy~\cite{wang2019symmetric}. These losses are all hand-designed based on various motivations. Instead we take a data-driven AutoML approach and search for a robust loss function. This draws upon meta-learning techniques but, differently from existing meta-robustness work, focuses on general-purpose white box loss discovery. Incidentally, we note that our resulting Taylor loss covers all six desiderata for noise-robust learning outlined in~\cite{song2020labelNoiseSurvey}. 

Finally, we mention one recent study \cite{yao2020searching} that also applied an AutoML approach to noisy label learning. In contrast to our approach, this method meta-learns a sample-selection technique to separate clean and noisy data, which must be conducted on a per-dataset basis. In our case, once trained, our loss is ready for plug-and-play deployment on diverse target problems with no further meta-training or dataset-specific optimization required (see Figure~\ref{fig:teaser}).

\textbf{Meta-learning and Loss Learning}\quad Meta-learning, also known as learning to learn, has been applied for a wide variety of purposes as summarized in~\cite{hospedales2020metaSurvey}. Of particular relevance is meta-learning of loss functions, which has been studied for various purposes including providing differentiable surrogates of non-differentiable objectives~\cite{huang2019addressing}, optimizing efficiency and asymptotic performance of learning \cite{jenni2018deep,bechtle2019meta,houthooft2018evolved,wu2018learning,gonzalez2019baikalLoss,gonzalez2020taylorGLO}, and improving robustness to train/test domain-shift \cite{balaji2018metareg,li2019feature}. We are particularly interested in learning \emph{white-box} losses for efficiency and improved task-transferability compared to neural network alternatives \cite{bechtle2019meta,houthooft2018evolved,balaji2018metareg,li2019feature}. Meta-learning of white-box learner components has been demonstrated for optimizers \cite{wichrowska2017learned}, activation functions \cite{ramachandran2017searching} and losses for accelerating conventional supervised learning \cite{gonzalez2019baikalLoss,gonzalez2020taylorGLO}. We are the first to demonstrate the value of automatic loss function discovery for general purpose label-noise robust learning.

\section{Method}
We aim to learn a loss function for multi-class classification problem that is robust to noisy labels in the training set. We consider the task of learning a loss function as a bilevel optimisation, where solutions generated by the upper objectives (in the outer loop) are conditioned on the response of the lower objectives (in the inner loop). In our loss function learning setting, the upper and lower problems are defined, respectively, as optimising the parameters of an adaptive loss function and training neural networks, $f_{\omega}$,  with the learned loss function. The upper level optimisation problem uses as a supervision signal a measure the average performance of models trained with a prospective loss function, measured across a variety of domains. The lower level optimisation problem consists of optimising a collection of models, each being trained to minimise the prospective loss function on a different domain that has been subjected to artificial label noise.

\begin{algorithm}[t]
\caption{Offline Taylor CMA-ES}
\label{general_algorithm}
\begin{algorithmic}[1]
\STATE {\bfseries Input: } $\mathcal{D}, F, \mu^{(0)}, \Sigma^{(0)}$
\STATE {\bfseries Output: } $p(\theta; \mu^{*}, \Sigma^{*})$
\STATE $t = 0$
\WHILE{not converged or reached max steps}
    \STATE Sample $\Theta = \{\theta_1, \theta_2, ..., \theta_n\} \sim p(\theta; \mu^{(t)}, \Sigma^{(t)})$ \COMMENT{Sample losses for exploration}
    \STATE $G = F\times\mathcal{D}\times\Theta$ 
        \COMMENT{Assign datasets and architectures to losses}
    \STATE $\vec s = \textup{zeros} \in \mathbb{R}^n$
    \FORALL{$(f^{(k)}, D_j, \theta_i) \in G$}
        \STATE $(D^{train}_j, D^{val}_j) = \textup{split}(D_j)$ \COMMENT{Construct train/val splits}
        \STATE $\omega^{*} = \argmin_\omega \mathcal{L}_{\theta_i}(f_\omega^{(k)}, D^{train}_j)$ \COMMENT{Train the network}
        \STATE $\vec s_{i} = \vec s_i + \frac{1}{|F||D|}\mathcal{M}(f_{\omega^{*}}^{(k)}, D^{val}_j)$ \COMMENT{Evaluate on validation data}
    \ENDFOR
    \STATE $(\mu^{(t+1)}, \Sigma^{(t+1)}) = \text{CMA-ES}(\mu^{(t)}, \Sigma^{(t)}, \Theta, \vec s)$ \COMMENT{Update $\mu$ and $\Sigma$ according to CMA-ES}
    \STATE $t = t + 1$
\ENDWHILE
\end{algorithmic}
\end{algorithm}

The prospective loss functions are represented by their parameters, $\theta$, which correspond to the coefficients of an $n$-th order polynomial. These polynomials can be viewed as a Taylor expansion of the ideal loss function. The bilevel optimisation problem is given by

\begin{align}
\max_{\theta} \, \mathbb{E}_{D, f} \lbrack \mathcal{M}(f_{\omega^{*}_D}, D^{val}) \rbrack \label{eq:bilevel_op} \\
s.t. \quad \omega^{*}_D = \argmin_{\omega} \, \mathcal{L}_{\theta}( f_\omega, D^{train}),\nonumber
\end{align}

\noindent where $\mathcal{M}(\cdot, \cdot)$ is a fitness function measuring network performance, $D$ is a random variable representing a domain, with $D^{val}$ and $D^{train}$ representing the validation and training sets respectively, and $f$ is a neural network parameterised by $\omega$. The performance of $f_{\omega^{*}_D}$, as measured by $\mathcal{M}$, reflects the quality of the supervision provided by the candidate loss function $\mathcal{L}_{\theta}$ on dataset $D$. During meta-learning, the training set and validation set are not identically distributed: the validation set contains clean labels, while the training set is assumed to have some form of label corruption. 
We use the Covariance Matrix Adaptation Evolutionary Strategy (CMA-ES)~\cite{hansen1996adapting} to solve the upper layer problem, and standard stochastic gradient-based optimisation approaches to solve the lower level problems. A general overview of our algorithm for solving the optimisation problem in Equations~\ref{eq:bilevel_op} is described in Algorithm~\ref{general_algorithm}.

\subsection{CMA-ES for Loss Function Learning}
We use CMA-ES to solve the upper optimisation problem, and any variant of stochastic gradient descent for the lower problem. CMA-ES finds a Gaussian distribution defined over the search space that places most of its mass on high quality solutions to the optimisation problem. One of the benefits of using CMA-ES is that this algorithm does not require the performance measurement to be differentiable, which means the learned loss function can be evaluated using informative metrics, such as accuracy. Each generation consists of a set, $\Theta$, of loss functions obtained by sampling multiple individuals from the parameter distribution, $p(\theta; \vec \mu, \Sigma) = \mathcal{N}(\vec \mu, \Sigma)$. Each of the individuals, $\theta_i \in \Theta$, is evaluated according to

\begin{align}
    \label{eq:fitness}
    \mathbb{E}_{D,f} \lbrack \mathcal{M}(f_{\omega^{*}_D}, D^{val}) \rbrack \approx \frac{1}{N} \sum_{j=1}^N \mathcal{M}(f_{\omega_j}^{(j)}, D_j^{val}) \\
    \textup{s.t.} \quad \omega_j = \argmin_{\omega} \, \mathcal{L}_{\theta_i}(f_\omega^{(j)}, D_j^{train}) \nonumber,
\end{align}

\noindent where $f^{(j)}_\omega$ and $D_j$ are different network architectures and datasets, respectively.

\subsection{Taylor Polynomial Representation}
The space of potential loss functions in which CMA-ES searches is a crucial design parameter. From a practical point of view, we must limit ourselves to a space that can be parameterised by a small number of values. However, this must be balanced with the ability to represent a wide enough variety of functions that a good solution can be found. Moreover, by selecting a small space with a small number of free parameters and well-understood nonlinear form, it becomes possible to transfer the learned loss to new problems without having to retrain the network. The function space that we choose is the Taylor series approximations of all $\beta$-smooth functions, $\mathcal{L}: \mathbb{R}^m \to \mathbb{R}$,
\begin{align}
    \mathcal{L}(\vec x) =& \sum_{n=0}^{\beta} \frac{1}{n!} \nabla^n \mathcal{L}(\vec x_0)^T(\vec x- \vec x_0)^{n}. \label{eq:taylor_raw}  
\end{align}
where each $\nabla^n \mathcal{L}(\vec x_0)$ is the $n$-th order gradient of $\mathcal{L}$ evaluated at a fixed point, $\vec x_0$. We make the simplifying assumption that the loss function should be class-wise separable. That is, each potential class is considered in isolation, and we learn a loss function that measures the divergence between a noisy binary label and the probability predicted by the network. We then sum over the different possible classes,
\begin{equation*}
    \mathcal{L}_\theta(\vec{\hat{y}}, \vec y) = \frac{1}{C}\sum_{i=1}^C \mathcal{L}_\theta^{(i)}(\vec{\hat{y}}_i, \vec y_i),
\end{equation*}
where $\vec{\hat{y}}$ and $\vec y$ are the vectors of predicted probabilities and (possibly noisy) ground-truth labels, respectively. The result of performing the simplification is that the loss function can be used in a variety of settings with different numbers of classes, and we can fix $m=2$. We found that $\beta=4$ is a good trade-off between modelling capacity and meta-training efficiency. Note that $\nabla^n \mathcal{L}(\vec x_0)$ does not depend on $\vec x$, meaning these values can be computed during a meta-training period before regular training commences. As such, we can reinterpret the task of meta-learning $\mathcal{L}$ as inferring $\theta = \{\nabla^n \mathcal{L}(\vec x_0)\}_{n=0}^\beta$. The resulting loss function is represented as 

\begin{align}
    \label{eq:taylorm}
    \mathcal{L}_{\theta}^{(i)}&(\vec{\hat{y}_i}, \vec{y_i}) = \vec{\theta}_2(\vec{\hat{y}}_i - \vec{\theta}_0) + \frac{1}{2}\vec{\theta}_3(\vec{\hat{y}}_i - \vec{\theta}_0)^2 \\ &+\frac{1}{6}\vec{\theta}_4(\vec{\hat{y}}_i - \vec{\theta}_0)^3 +\frac{1}{24}\vec{\theta}_5(\vec{\hat{y}}_i-\vec{\theta}_0)^{4} \nonumber \\
    & + \vec{\theta}_6(\vec{\hat{y}}_i - \vec{\theta}_0)(\vec{y}_i - \vec{\theta}_1) \nonumber \\
    &+ \frac{1}{2}\vec{\theta}_7(\vec{\hat{y}}_i-\vec{\theta}_0)(\vec{y}_i -\vec{\theta}_1)^2  
    + \frac{1}{2} \vec{\theta}_8(\vec{\hat{y}}_i - \vec{\theta}_0)^2(\vec{y}_i - \vec{\theta}_1)  \nonumber \\
    & + \frac{1}{6}\vec{\theta}_9(\vec{\hat{y}}_i - \vec{\theta}_0)^3(\vec{y}_i - \vec{\theta}_1) 
    + \frac{1}{6}\vec{\theta}_{10}(\vec{\hat{y}}_i - \vec{\theta}_0)(\vec{y}_i - \vec{\theta}_1)^3 \nonumber\\
    &+ \frac{1}{4}\vec{\theta}_{11}(\vec{\hat{y}}_i - \vec{\theta}_0)^2(\vec{y}_i - \vec{\theta}_1)^2. \nonumber
\end{align}

The fixed point where the gradients are evaluated is also left as a learned parameter, $(\vec{\theta}_{0}, \vec{\theta}_{1})$. 
Note that we have omitted terms where $\vec{\hat{y}}$ does not appear, as these do not impact the solution of the optimisation problem. In total there are only 12 parameters to fit, which is considerably smaller than the number of parameters found in a typical neural network paramaterized loss function~\cite{li2019feature,bechtle2019meta,kirsch2020Improving}.

\subsection{Generalisation Across Architectures}
To enable the learned loss function to generalise to different architectures, we extend the strategy domain randomisation \cite{tobin2017domain} to introduce the idea of evaluating the expected performance across a range of architectures during meta-learning. Specifically, we use a set, $F$, of $m$ architectures containing a variety of common neural network designs. The total population for evolutionary optimisation is then given by the Cartesian product $F \times \Theta$. The fitness function can then be computed as shown in Equation~\ref{eq:fitness}, where a mean is taken over all of the different architectures trained with the same loss.

\subsection{Generalisation Across Datasets}
We also explore another method for improving the generality of the loss function. To enable a loss function to be applied to an unseen dataset, the loss function should be exposed to several datasets during training so as not to overfit to the prediction distributions encountered for a specific machine learning problem. In our method a loss function sampled from the current target distribution is deployed to train several models with the same architecture and initial weights, but on different datasets. Similarly to architecture generalisation, we use a set of datasets ,$\mathcal{D}$, and take the Cartesian product, $\mathcal{D} \times \Theta$, to generate a population to be evaluated. The performance of the loss functions is evaluated by the mean performance of all the networks on their corresponding datasets.

In principle, one could perform both dataset and architecture randomisation simultaneously. However, due to the implied three-way Cartesian product, we found this computationally infeasible.

\subsection{Normalisation \label{sec:minmax}}
We make use of a normalisation approach to prevent the learned loss functions from exhibiting an arbitrary output range,
\begin{equation}
\hat{f} = \eta \frac{f-f_{min}}{f_{max}-f_{min}}, \label{eq:min_max}
\end{equation}
where $f_{min}$ and $f_{max}$ denotes the minimum and maximum and $\eta$ is a hyperparameter deciding the dynamic range of the loss function. Both $f_{min}$ and $f_{max}$ are easily approximated by sampling random points satisfying $\{(\vec{\hat{y}}, \vec{y})| \vec{\hat{y}}_i \geq 0, \sum_{i}\vec{\hat{y}}_i = 1; \vec{y}_i \in \{0,1\}, \sum_{i}\vec{y} = 1\}$, which defines the domain of the loss function.
\section{Experiments}
In this section we evaluate our learned loss function on various noisy label learning tasks. In particular, we aim to answer three questions: (Q1) Can we learn a robust loss function that generalises across different datasets and architectures? (Q2) How well does our learned loss function generalise across different noise levels? (Q3) Can our learned loss function scale to larger scale real-world noisy-label tasks? 

\textbf{Datasets}\quad We use seven datasets in our experiments:  MNIST~\cite{lecun-mnisthandwrittendigit-2010}, CIFAR-10, CIFAR-100~\cite{krizhevsky2009learning}, KMNIST~\cite{clanuwat2018deep}, USPS~\cite{hull1994database}, FashionMNIST~\cite{xiao2017fashion} and Clothing1M~\cite{xiao2015learning}. Clothing1M is a dataset containing 1 million clothing images in 14 classes: T-shirt, Shirt, Knitwear, Chiffon, Sweater, Hoodie, Windbreaker, Jacket, Down Coat, Suit, Shawl, Dress, Vest, and Underwear. The images are collected from shopping websites and the labels are generated from the text surrounding images, thus providing a realistic noisy label setting. 

\begin{figure}[t]
\begin{minipage}[b]{.47\linewidth}
  \centering
 \centerline{\epsfig{figure={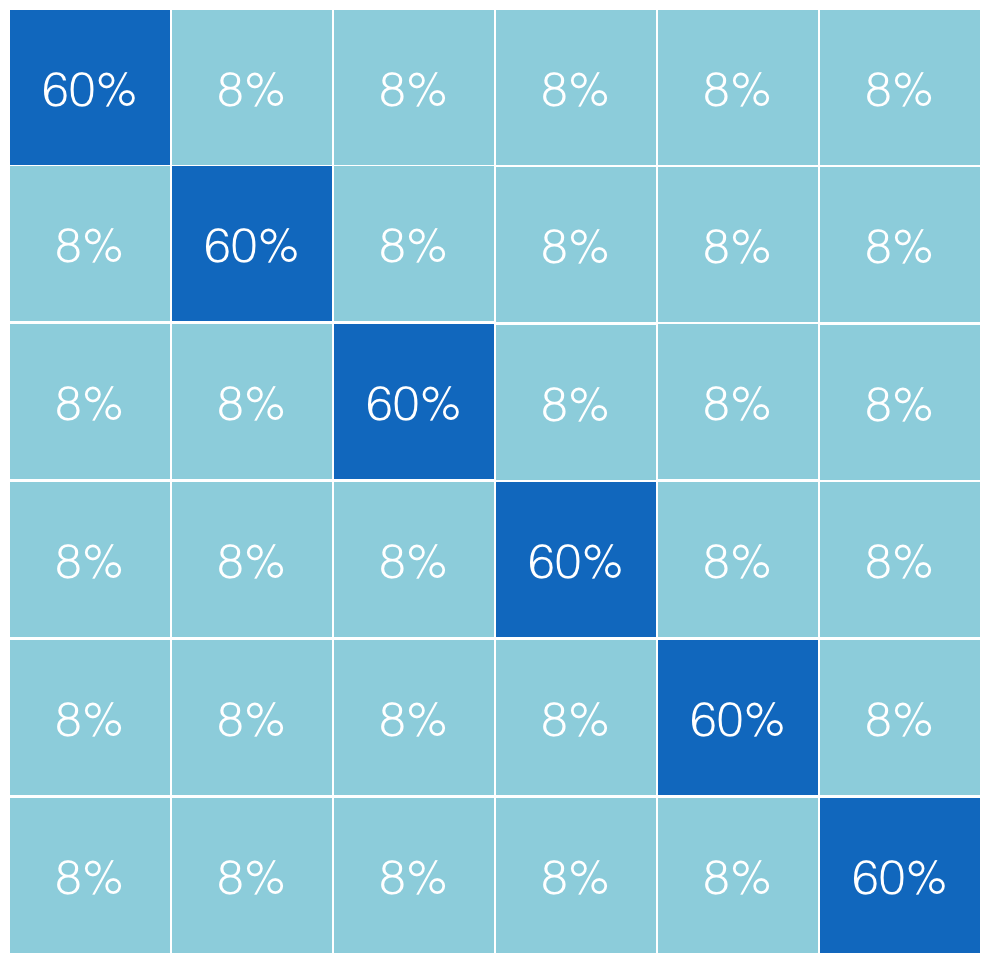},width=4cm}}
\end{minipage}
\hfill
\begin{minipage}[b]{.47\linewidth}
  \centering
 \centerline{\epsfig{figure={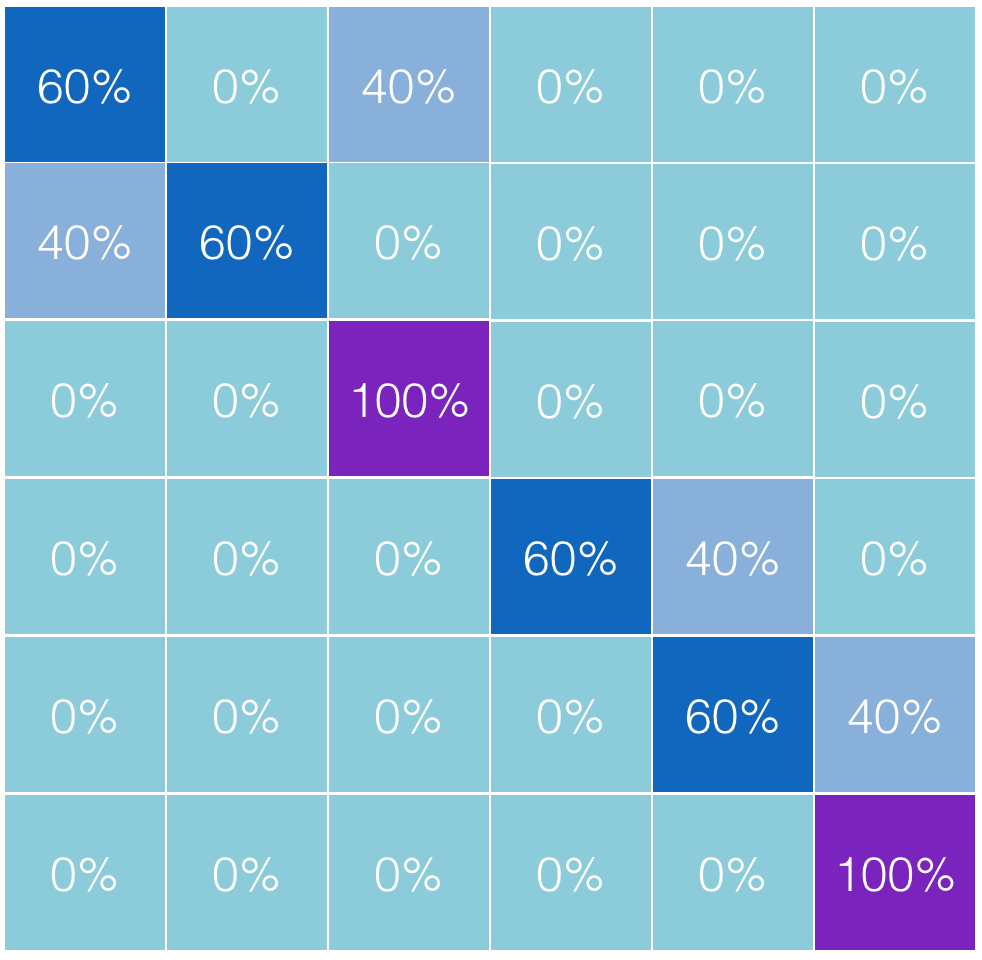},width=4cm}}
\end{minipage}
\caption{Example noise matrix in the symmetric  (left) and asymmetric  (right) conditions for 6 classes and a noise ratio of 40\%.} 
\label{fig:noise_matrix}
\end{figure}

\textbf{Noise types}\quad For loss learning, we consider simulating two types of noise, symmetric noise and asymmetric noise (pair-flip noise). Symmetric noisy labels are generated by uniformly flipping from the positive label to a negative one, while asymmetric noisy labels are produced to simulate the more realistic scenario where particular pairs of categories are more easily confused than others. {For example, in the case of MNIST it is conceivable that label noise could manifest in such a way that a 7 is more likely to mislabelled as a 1 than it is a 6, or a 3 mislabelled as an 8 than a 4. We give an example of both symmetric and asymmetric label noise transition matrices in Figure \ref{fig:noise_matrix}.}

\textbf{Architectures}\quad We train and evaluate our learned loss with a range of neural networks from very shallow ones, including 2-layer MLP, 3-layer MLP, and 4-layer CNN, to deeper ones, such as VGG-11~\cite{simonyan2014very} and Resnet-18~\cite{he2016deep}. We also use the medium-size architecture considered in~\cite{wei2020combating}, which we term JoCoR-Net (see  supplemental material for details). For a fair comparison, we train 2-layer MLP, 3-layer MLP, and 4-layer CNN with SGD optimiser and set the learning rate to $0.01$ and momentum to $0.9$. For the training of JoCor-Net, we apply the Adam optimiser~\cite{kingma2014adam} and the learning rate is set to $0.001$. When training Resnet-18 and VGG-11, we follow the training protocol in~\cite{zhang2019lookahead}.

\textbf{Taylor Polynomial Order Selection}\quad We perform a preliminary experiment to select the order of the Taylor loss function. We train a linear classifier in the inner loop of the  dataset randomization algorithm (on MNIST, KMNIST, and CIFAR-10), and evaluate performance for polynomial orders 2, 3, 4 and 5. From the results in Figure~\ref{fig:shared}(left), we can see that the impact of the specific polynomial order is small compared to the impact of loss learning overall. Nevertheless, we pick order 4 for the subsequent experiments, as this was the hyperparameter that achieved the best performance.

\textbf{Competitors}\quad We compare our learned loss functions with the standard cross-entropy (CE) baseline, as well as several strong alternative losses hand-designed for label-noise robustness: \textbf{MAE:} Mean Absolute Error was theoretically shown to be robust in \cite{ghosh2017robust}. \textbf{GCE:} \cite{zhang2018generalized} analysed MAE as hard to train, and proposed  generalised cross-entropy to provide the best of CE and MAE; \textbf{FW:}~\cite{patrini2017making} iteratively estimates the label noise transfer matrix, and trains the model corrected by the label noise estimate; \textbf{SCE:}~\cite{wang2019symmetric} argued that symmetrizing cross-entropy by adding reverse cross-entropy (RCE) improves label-noise robustness; \textbf{Bootstrap:} A classic method of replacing the noisy labels in training by the convex combination of the prediction and the given labels ~\cite{reed2014training}. \textbf{LSR:} Label-smoothing is an effective general purpose regularizer \cite{pereyra2017regularizing,szegedy2016rethinking,muller2019labelSmooth} whose properties in promoting noise robustness have been studied \cite{wang2019symmetric}.

\subsection{Training a general-purpose robust loss function \label{se:generality}}

\begin{figure*}[t]
\centering
\begin{minipage}[b]{.33\textwidth}
\noindent\includegraphics[width=1.1\linewidth]{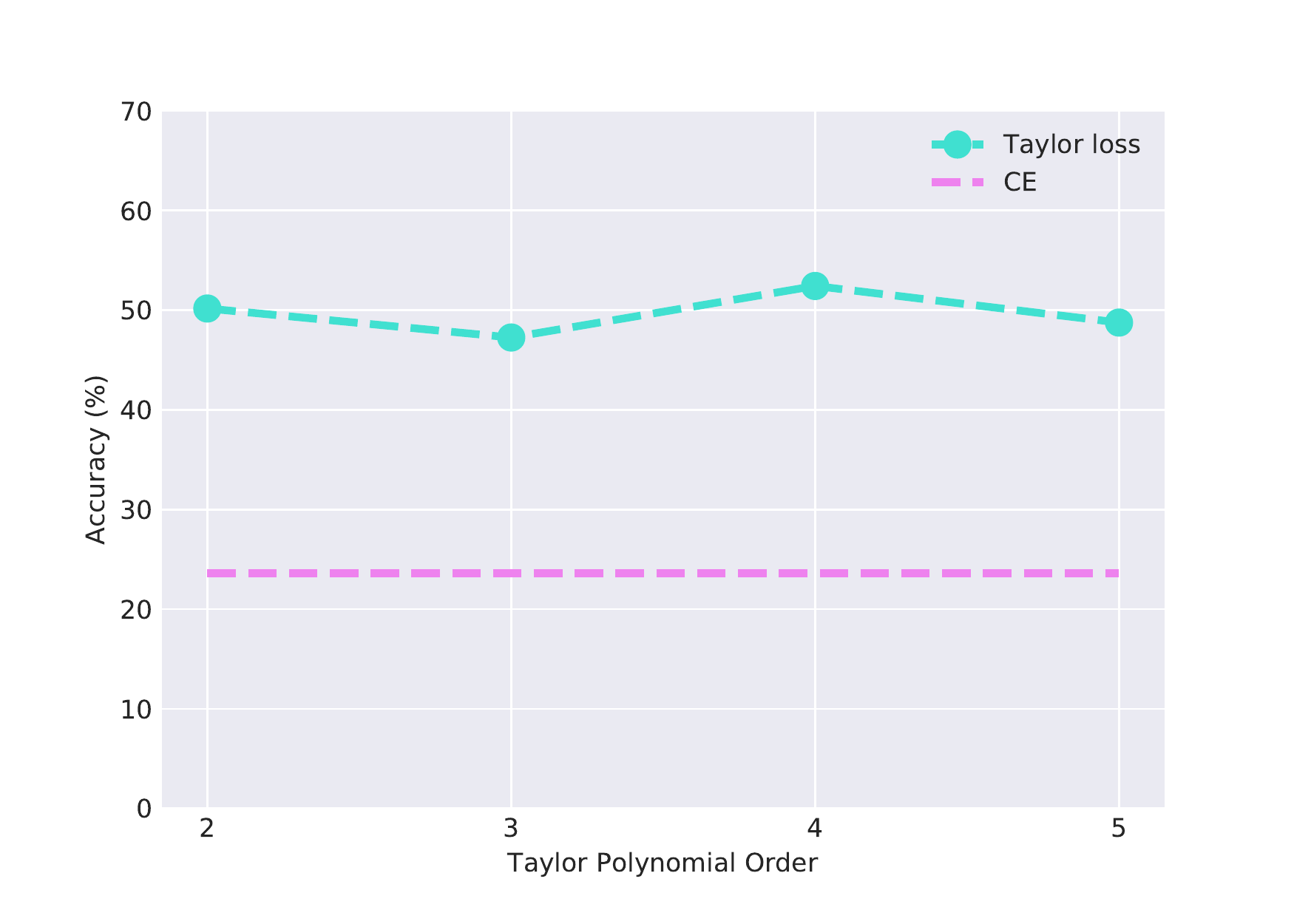}
\end{minipage}
\begin{minipage}[b]{.33\textwidth}
\noindent\includegraphics[width=1.1\textwidth]{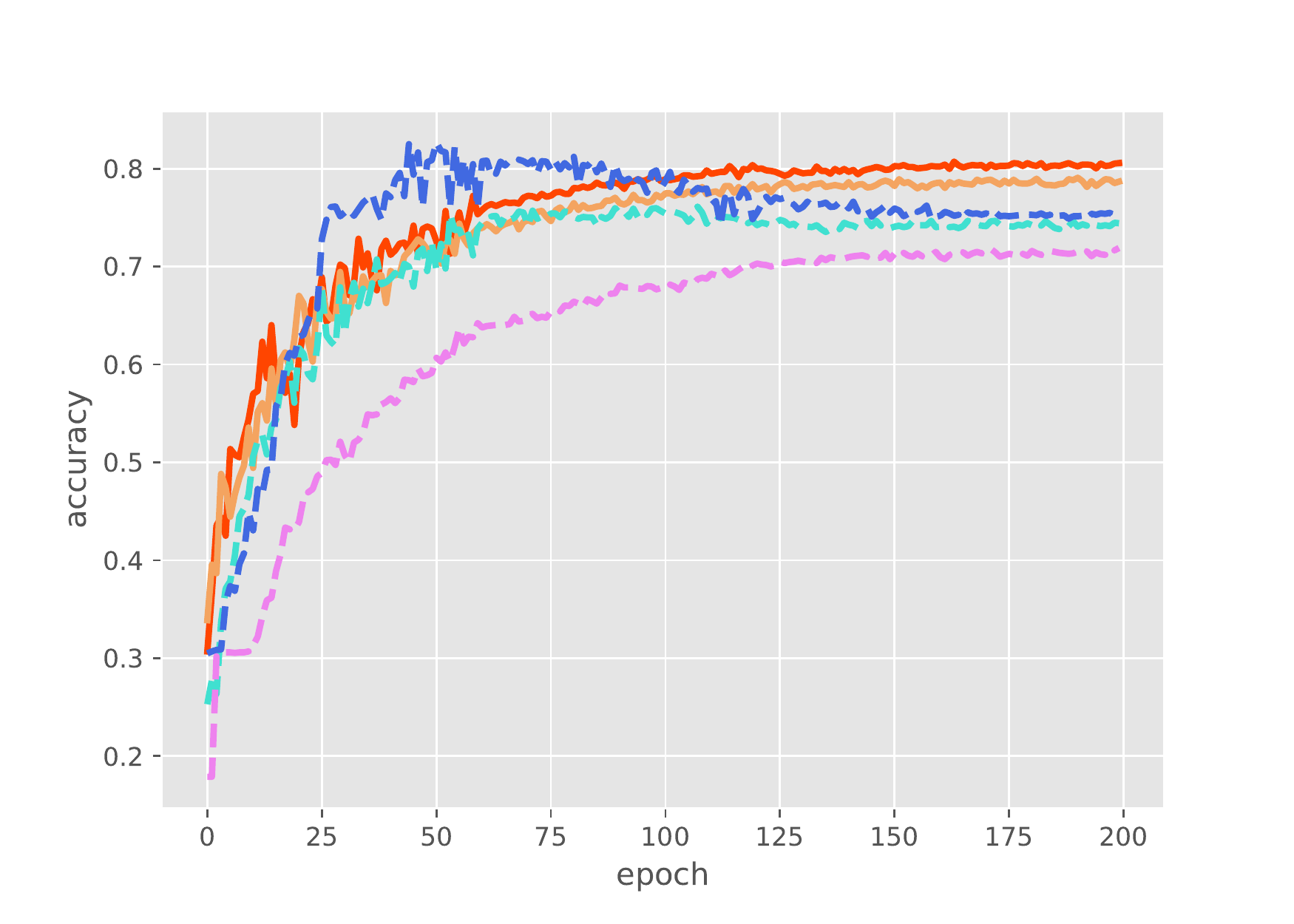}
\end{minipage}
\begin{minipage}[b]{.33\textwidth}
\noindent\includegraphics[width=1.1\textwidth]{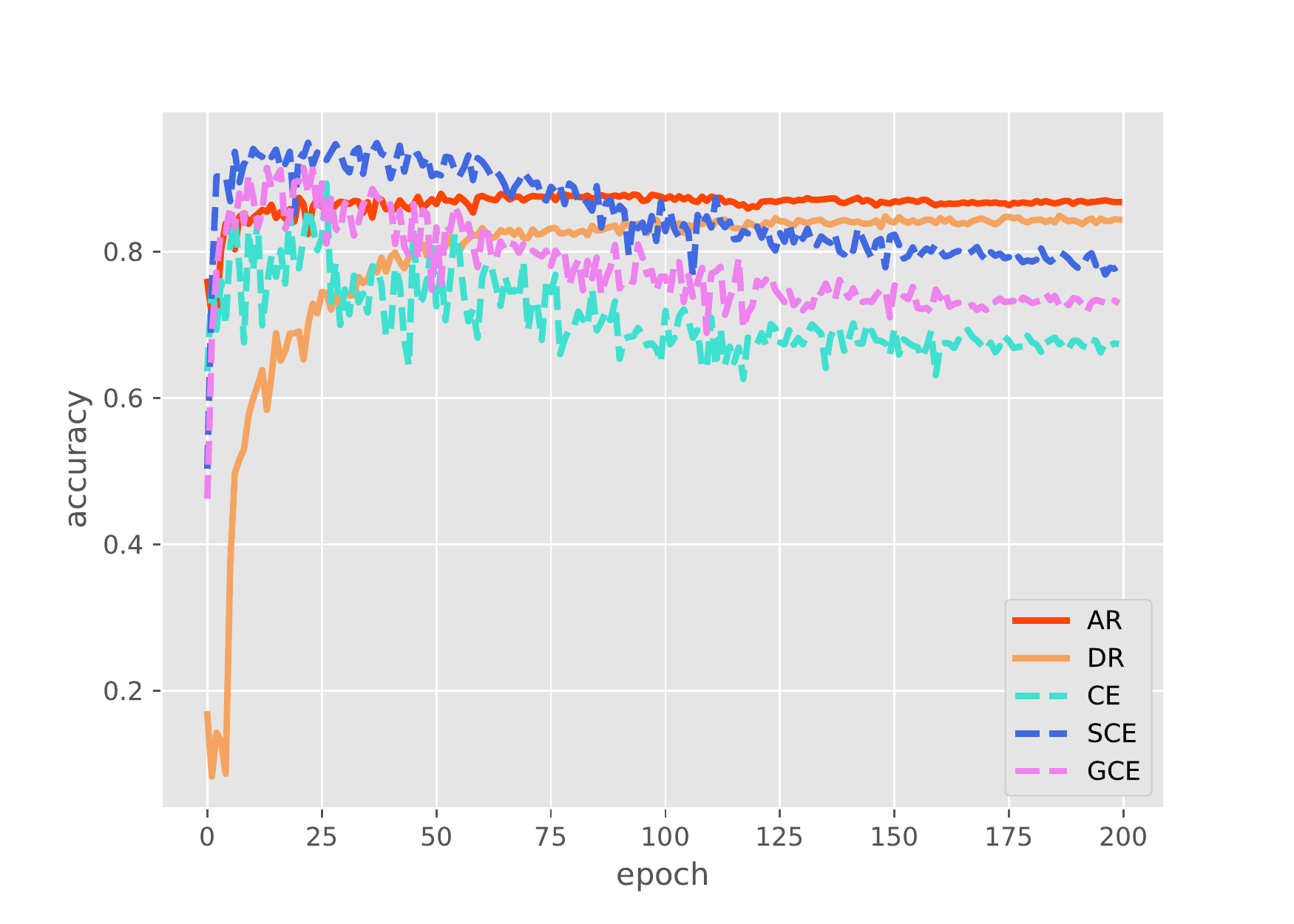}
\end{minipage}
    \centering
    \caption{Left: A preliminary experiment on hyperparameter selection. The performance of a linear model trained by the Taylor loss function with different orders vs training with cross-entropy (CE). Middle/Right: Example learning curves of test accuracy vs iterations when using different robust losses. Middle: USPS/VGG-11/80\% symmetric noise. Right: USPS/ResNet-18/40\% asymmetric noise. }\label{fig:shared}
\end{figure*}
\textbf{Experimental setup}\quad We consider two domain generalisation protocols for training a general purpose loss function, namely architecture and dataset generalisation. In architecture generalisation, we build a pool of training architectures including 2-layer MLP, 3-layer MLP, and 4-layer CNN and solely use MNIST as the training set. In dataset generalisation, we solely use the 4-layer CNN as the backbone build a dataset pool from  MNIST, KMNIST, and CIFAR-10. We also consider two label-noise conditions during meta-learning: symmetric label noise with 80\% noise, and asymmetric label noise with 40\% noise. Losses are trained under each domain generalisation protocol, and each noise distribution, using the normalisation trick introduced in Section~\ref{sec:minmax}. 
It is important to note that we are not restricted to deploying our loss on the same tasks used during training. After loss function learning we can deploy the losses to train fresh models from scratch on a fresh suite of evaluation tasks unseen during training. For evaluation, we compare the accuracy at convergence, and summarise via the average ranks of each methods across different datasets and architectures~\cite{demvsar2006statistical}. 

\begin{table*}[t]
\centering
\caption{Accuracy (\%) of different robust learning strategies. 80\% symmetric noise condition. Our loss trained under architecture randomization (AR) and dataset randomization (DR) conditions has the best average rank. Grey columns indicate datasets seen during DR training. White columns are totally novel datasets.}
\label{tab:_gen}
\resizebox{\textwidth}{!}{
\begin{tabular}{ laaacccacccc}
\toprule
Architecture type & 2layer MLP & 4layer CNN & VGG11   & VGG11    & VGG11        & VGG11 & Resnet18 & Resnet18 & Resnet18     & Resnet18 &Avg.Rank \\
dataset           & MNIST      & MNIST      & Cifar10 & Cifar100 & FashionMNIST & USPS  & Cifar10  & Cifar100 & FashionMNIST & USPS     &        \\
\midrule
CE                & 22.10$\pm$0.68         & 28.48$\pm$0.35        & 18.38$\pm$0.21   &  4.25$\pm$0.28  & 20.55 $\pm$ 0.93 & 51.42 $\pm$ 0.94 &  18.44$\pm$0.34       &  8.86$\pm$0.10   & 21.92 $\pm$ 0.74 & 57.05 $\pm$ 0.42       & 6.4\\
GCE               & 40.57$\pm$0.43         &  9.80$\pm$0.58        & 16.56$\pm$0.54   &  1.04$\pm$0.47  & 25.10 $\pm$ 0.68 & 63.45 $\pm$ 0.86 &  31.69$\pm$0.36       & 11.98$\pm$0.18   & 42.62 $\pm$ 0.89 & \bfs{79.52 $\pm$ 0.63} & 5.1 \\
SCE               & 31.23$\pm$0.70         & 28.53$\pm$1.02        & 28.61$\pm$0.64   &  2.31$\pm$0.80  & 36.64 $\pm$ 0.59 & 63.68 $\pm$ 0.56 &  \bfs{45.34$\pm$0.40} &  8.16$\pm$0.07   & 59.93 $\pm$ 0.75 & 58.35 $\pm$ 0.76       & 4.6 \\
FW                & 54.01$\pm$0.89         & 80.34$\pm$1.21        & 16.97$\pm$0.44   &  1.41$\pm$0.07  & 22.57 $\pm$ 0.76 & 53.66 $\pm$ 0.40 &  10.15$\pm$0.68       &  1.16$\pm$0.04   & 13.18 $\pm$ 0.35 & 42.80 $\pm$ 0.77       & 6.5 \\
Bootstrap         & 23.46$\pm$1.31         & 28.78$\pm$1.03        & 17.58$\pm$0.82   &  4.18$\pm$0.72  & 20.40 $\pm$ 0.31 & 64.58 $\pm$ 0.21 &  12.10$\pm$0.32       &  8.67$\pm$0.61   & 22.36 $\pm$ 1.76 & 72.17 $\pm$ 1.24       & 5.7 \\
MAE               & \bfs{85.40$\pm$3.39}   & 78.70$\pm$11.49       & 14.20$\pm$0.42   &  1.01$\pm$0.11  & 63.40 $\pm$ 0.16 & 30.94 $\pm$ 0.35 &  22.95$\pm$1.25       &  0.82$\pm$0.17   & 68.20 $\pm$ 1.87 & 37.17 $\pm$ 0.93       & 6.0 \\
Label-smooth      & 24.31$\pm$1.25         & 27.02$\pm$0.48        & 17.74$\pm$0.46   &  4.47$\pm$0.12  & 21.19 $\pm$ 0.39 & 54.26 $\pm$ 0.19 &  17.67$\pm$0.35       &  7.66$\pm$1.52   & 20.99 $\pm$ 0.83 & 59.94 $\pm$ 0.54       & 6.3 \\
\midrule
Taylor Loss (AR)  & 34.27$\pm$0.34         & 37.08$\pm$0.44        & \bfs{41.36$\pm$0.47}  &  \bfs{5.63$\pm$0.24} & \bfs{70.16 $\pm$ 0.87} & \bfs{78.71 $\pm$ 0.90} & 29.50$\pm$0.30 & \bfs{14.94$\pm$0.26}  & 71.96 $\pm$ 0.89       & 68.80 $\pm$ 0.92 & 2.4 \\
Taylor Loss (DR)  & 48.57$\pm$0.11         & \bfs{85.31$\pm$0.12}  & 31.12$\pm$0.23        &  5.04$\pm$0.14       & 67.29 $\pm$ 1.01       & 77.34 $\pm$ 1.34       & 35.23$\pm$0.23 & 13.36$\pm$0.63        & \bfs{71.97 $\pm$ 0.87} & 70.17 $\pm$ 0.64 & \bfs{2.0} \\
\bottomrule
\end{tabular}
}
\end{table*}

\begin{table*}[t]
\centering
\caption{Accuracy (\%) of different robust learning strategies. 40\% asymmetric noise condition. Our loss trained under architecture randomization (AR) and dataset randomization (DR) conditions has the best average rank. Grey columns indicate datasets seen during DR training. White columns are totally novel datasets.}
\label{tab:pairnoise_gen}
\resizebox{\textwidth}{!}{
\begin{tabular}{ laaacccacccc}
\toprule
Architecture type & 2layer MLP & 4layer CNN & VGG11   & VGG11    & VGG11           & VGG11  & Resnet18 & Resnet18 & Resnet18     & Resnet18 & Avg.Rank\\
dataset           &  MNIST     & MNIST      & Cifar10 & Cifar100 & FashionMNIST    & USPS  & Cifar10  & Cifar100 & FashionMNIST  & USPS     &         \\
\midrule
CE            & 78.73$\pm$1.16       & 84.01$\pm$0.34       & 56.43$\pm$0.12       & 30.20$\pm$0.18       & 50.34 $\pm$ 1.23 & 77.74 $\pm$ 0.74 & 58.69$\pm$0.43       & 44.14$\pm$0.15       & 58.68 $\pm$ 0.63 & 73.84 $\pm$ 0.85 & 5.0\\
GCE           & 81.94$\pm$1.22       &  9.80$\pm$0.10       & 56.42$\pm$0.54       & 22.39$\pm$0.35       & 53.57 $\pm$ 0.47 & 78.72 $\pm$ 0.72 & 57.90$\pm$0.31       & 40.76$\pm$0.24       & 58.51 $\pm$ 0.70 & 80.77 $\pm$ 0.35 & 5.3\\
SCE           & 79.87$\pm$0.78       & 84.09$\pm$0.62       & 78.23$\pm$0.55       & 25.33$\pm$0.73       & 64.47 $\pm$ 0.97 & 85.50 $\pm$ 0.43 & 63.22$\pm$0.22       & 40.90$\pm$0.37       & 59.63 $\pm$ 0.96 & 81.57 $\pm$ 0.17 & 3.2\\
FW            & 90.14$\pm$0.67       & 69.98$\pm$0.49       & 54.42$\pm$0.79       &  5.21$\pm$0.39       & 45.18 $\pm$ 0.84 & 76.41 $\pm$ 0.81 & 48.40$\pm$0.08       &  3.83$\pm$0.23       & 49.46 $\pm$ 0.73 & 46.04 $\pm$ 0.18 & 7.6\\
Bootstrap     & 78.31$\pm$2.34       & 83.68$\pm$1.27       & 57.69$\pm$0.11       & \bfs{31.07$\pm$1.09} & 53.23 $\pm$ 1.53 & 77.81 $\pm$ 0.61 & 57.69$\pm$0.76       & \bfs{45.78$\pm$0.15} & 54.60 $\pm$ 0.85 & 75.67 $\pm$ 0.56 & 5.0\\
MAE           & 71.61$\pm$4.50       & 69.91$\pm$0.49       & 49.06$\pm$0.22       &  0.96$\pm$0.10       & 49.02 $\pm$ 0.27 & 62.38 $\pm$ 0.89 & 55.67$\pm$3.05       &  1.02$\pm$0.14       & 56.31 $\pm$ 1.21 & 70.05 $\pm$ 0.35 & 8.2\\
Label-smooth  & 59.66$\pm$1.16       & 68.14$\pm$0.61       & 57.76$\pm$0.37       & 20.64$\pm$0.18       & 51.12 $\pm$ 1.03 & 77.49 $\pm$ 0.11 & 59.69$\pm$0.36       & 39.92$\pm$0.49       & 57.53 $\pm$ 0.73 & 78.97 $\pm$ 0.46 & 6.1\\
\midrule
Taylor Loss (AR)     & \bfs{97.16$\pm$0.20} & \bfs{96.88$\pm$0.66} & 74.30$\pm$0.20       & 22.50$\pm$0.33       & \bfs{87.23 $\pm$ 1.22} & \bfs{90.67$\pm$1.21} & \bfs{86.70$\pm$0.12} & 44.47$\pm$0.48    & \bfs{89.24 $\pm$ 0.25} & \bfs{91.17 $\pm$ 0.25} & \bfs{1.6}     \\
Taylor Loss (DR)     & 85.77$\pm$0.33       & 93.47$\pm$0.28       & \bfs{79.09$\pm$0.51} & 18.30$\pm$0.27       & 81.18 $\pm$ 0.80       & 89.78$\pm$0.46       & 68.88$\pm$0.41       & 31.47$\pm$0.65    & 88.22 $\pm$ 0.97       & 89.59 $\pm$ 1.05       & 3.0\\
\bottomrule
\end{tabular}
}
\end{table*}

\textbf{Benchmark Results}\quad 
The results for symmetric and asymmetric noise are shown in Table~\ref{tab:_gen} and \ref{tab:pairnoise_gen} respectively. From the results, we can see that our learned losses perform favourably compared to hand-designed alternatives across a variety of benchmarks, with our learned loss providing a higher average rank than competitors in both experiments. However, there is no clear winner between architecture (AR) and dataset (DR) condition for meta-learning. We conjecture that best performance would be obtained by performing these simultaneously, but as this experiment is computationally costly, we leave this to future work. Note that during deployment, all methods have a similar computational cost, except for FW which requires training the network twice for noise estimation. 

\textbf{Analysis of Learning Curves}\quad The plots in Figure~\ref{fig:shared}(right) compares the learning curves of test accuracy for USPS/VGG-11 and USPS/ResNet-18 with 80\% symmetric and 40\% asymmetric noise respectively. We can see that while some alternative losses have early peaks, they all overfit after continued training. It is important to note that because we target the situation where we do not have a clean validation set for the target domain to drive model selection, one cannot rely on early-stopping cherry pick a good iteration. Therefore it's important that a robust loss has longer-term good performance, and on this metric our Taylor-AR and Taylor-DR are the clear winners.

\cut{
\begin{figure}[t]
\begin{minipage}[b]{.47\linewidth}
  \centering
 \centerline{\epsfig{figure={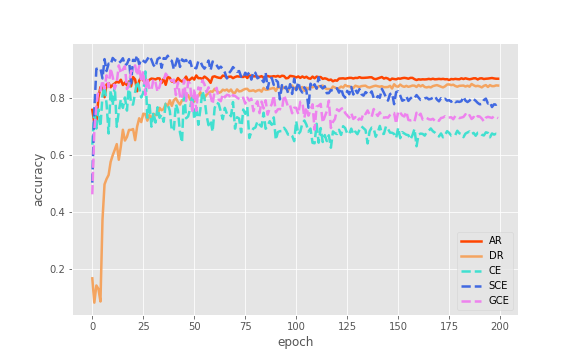},width=5cm}}
\end{minipage}
\hfill
\begin{minipage}[b]{.47\linewidth}
  \centering
 \centerline{\epsfig{figure={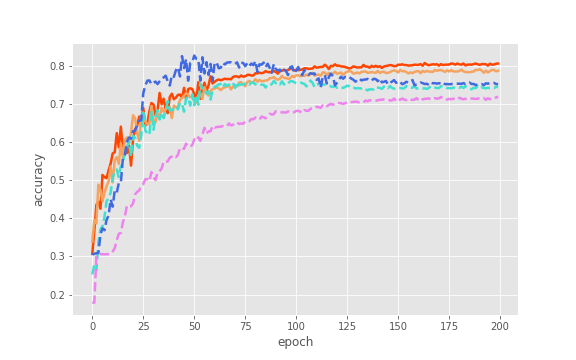},width=5cm}}
\end{minipage}
\caption{The test accuracy alongside with training epochs. Right: Resent18 trained on USPS with 40 \% asymmetric noise. Left: VGG11 trained on USPS with 80 \% symmetric noise } 
\label{fig:acc_curve}
\end{figure}
}

\textbf{Real-world Clothing1M results}\quad The previous experiment reported performance of the learned model after training on manually corrupted labels. In this section, we follow the setting for Resent-18 described in~\cite{wei2020combating} to apply our learned loss to the real-world Clothing1M noisy-label benchmark. As a real-world noisy-label problem we apply our model from the asymmetric-40\% condition above. Note that neither Clothing1M, nor ResNet-18 were seen during meta-learning, above. We train with Adam optimiser with learning rate $8\times10^{-4}$, $5\times10^{-4}$, $5\times10^{-5}$ for 5 epochs each in a sequence. We report the mean accuracy of each model after ten trials in Table~\ref{tab:clothing1mNew}. Among the competitors, JoCoR is the state art method in the broader range of noise robust learners. It uses a complex co-distillation scheme with multiple network branches, while the other listed competitors and ours are simple plug-in robust losses applied to vanilla ResNet training. Nevertheless, our method obtains the highest performance. 
\cut{
\begin{table}[t]
\centering
\caption{Test accuracy ($\%$) of robust learners on Clothing1M with ResNet18. $^*$JoCoR is a multi-network co-distillation training framework. The others are simple plug-in robust losses.}
\label{tab:clothing1m}
\resizebox{1\columnwidth}{!}{
\begin{tabular}{ c c c c c c }
\toprule
\midrule
CE      & Bootstrap  & GCE    & FW     & SCE     & JoCoR$^*$   \\
66.88   & 67.28      & 66.63  & 68.33  & 67.63   & 69.79 \\ 
\midrule
& Talyor (AR-A40) & Taylor (DR-A40) & Taylor (AR-S80) & Taylor (DR-S80)&\\
& 69.14             & \textbf{70.09}     & 68.85             & 69.34 &\\
\bottomrule
\end{tabular}
}
\end{table}
}

\cut{\begin{table*}[t]
\centering
\caption{Test accuracy ($\%$) of robust learners on Clothing1M with ResNet18. $^*$JoCoR is a multi-network co-distillation training framework. The others are simple plug-in robust losses.}
\label{tab:clothing1m}
\resizebox{\textwidth}{!}{
\begin{tabular}{ c c c c c c c c c c }
\toprule
CE      & Bootstrap  & GCE    & FW     & SCE     & JoCoR$^*$       & Talyor (AR-A40) & Taylor (DR-A40) & Taylor (AR-S80) & Taylor (DR-S80)\\
\midrule
66.88   & 67.28      & 66.63  & 68.33  & 67.63   & 69.79 & 69.14   & \textbf{70.09}     & 68.85             & 69.34 \\
\bottomrule
\end{tabular}
}
\end{table*}}

\begin{table}[t]
\centering
\caption{Test accuracy ($\%$) of robust learners on Clothing1M with ResNet18. $^*$JoCoR is a multi-network co-distillation training framework. The others are simple plug-in robust losses.}\label{tab:clothing1mNew}
\begin{tabular}{lc}
\toprule
Method & Accuracy \\
\midrule
CE              & 66.88                           \\ 
Bootstrap       & 67.28                           \\ 
GCE             & 66.63                           \\ 
FW              & 68.33                           \\ 
SCE             & 67.63                           \\ 
JoCoR$^*$       & 69.79                           \\
\midrule
Talyor (AR-A40) & 69.14                           \\ 
Taylor (DR-A40) & \textbf{70.09} \\ 
Taylor (AR-S80) & 68.85                           \\ 
Taylor (DR-S80) & 69.34                           \\
\bottomrule
\end{tabular}
\end{table}

\begin{figure*}[t]
\begin{minipage}[b]{.32\linewidth}
\centering
\centerline{\epsfig{figure={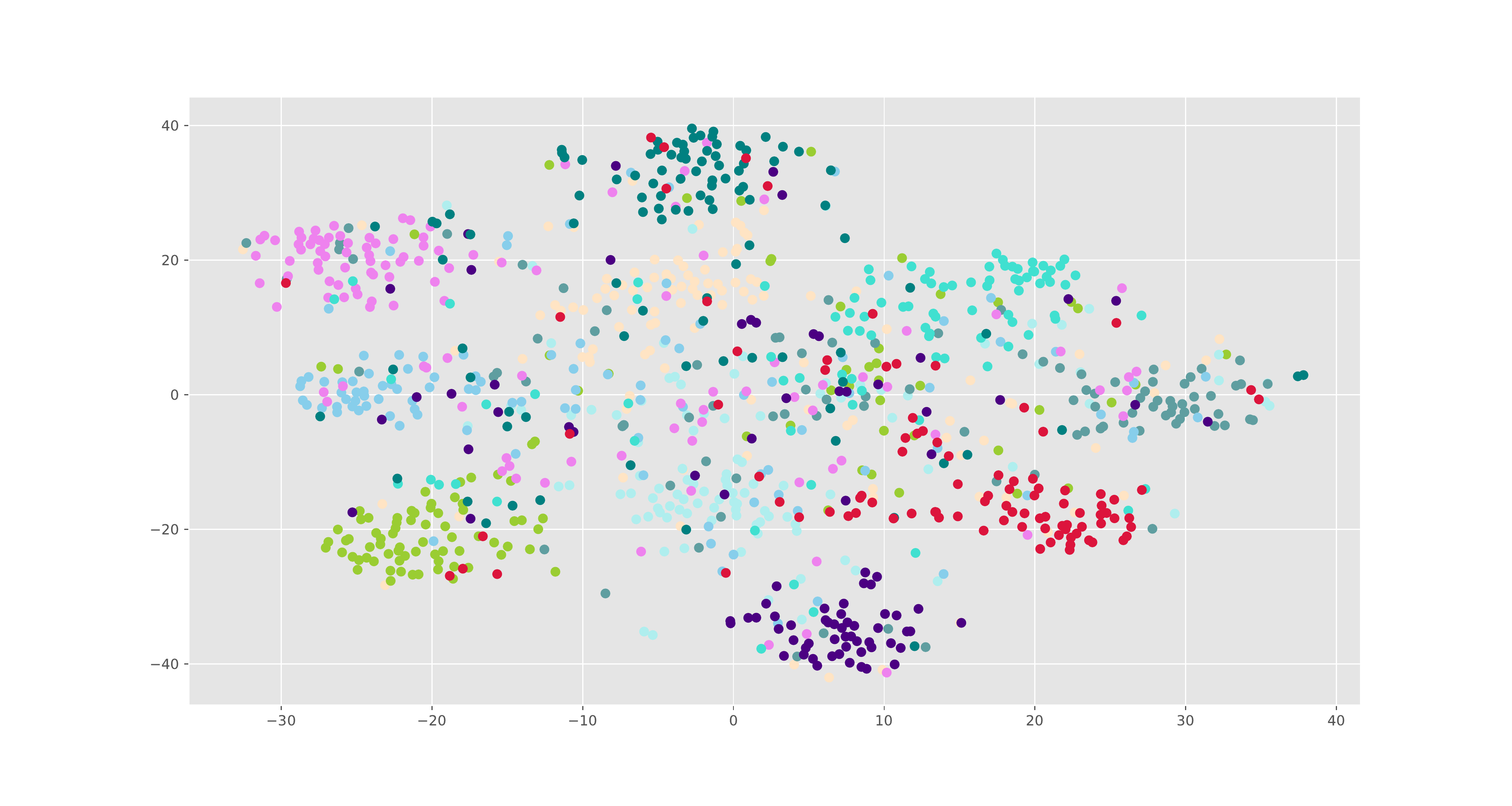},width=6.4cm}}
\end{minipage}
\hfill
\begin{minipage}[b]{.32\linewidth}
\centering
\centerline{\epsfig{figure={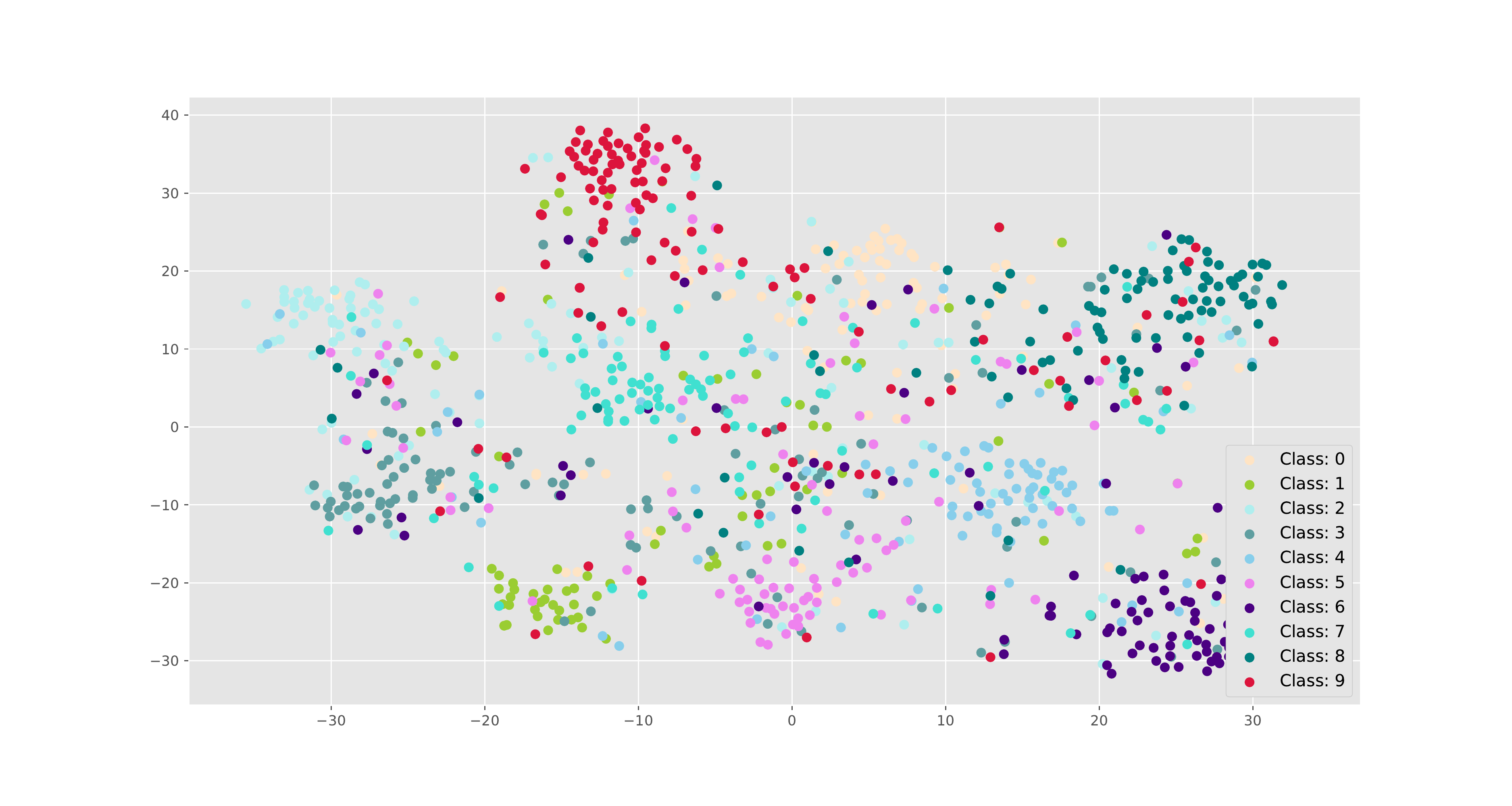},width=6.4cm}}
\end{minipage}
\hfill
\begin{minipage}[b]{.32\linewidth}
\centering
\centerline{\epsfig{figure={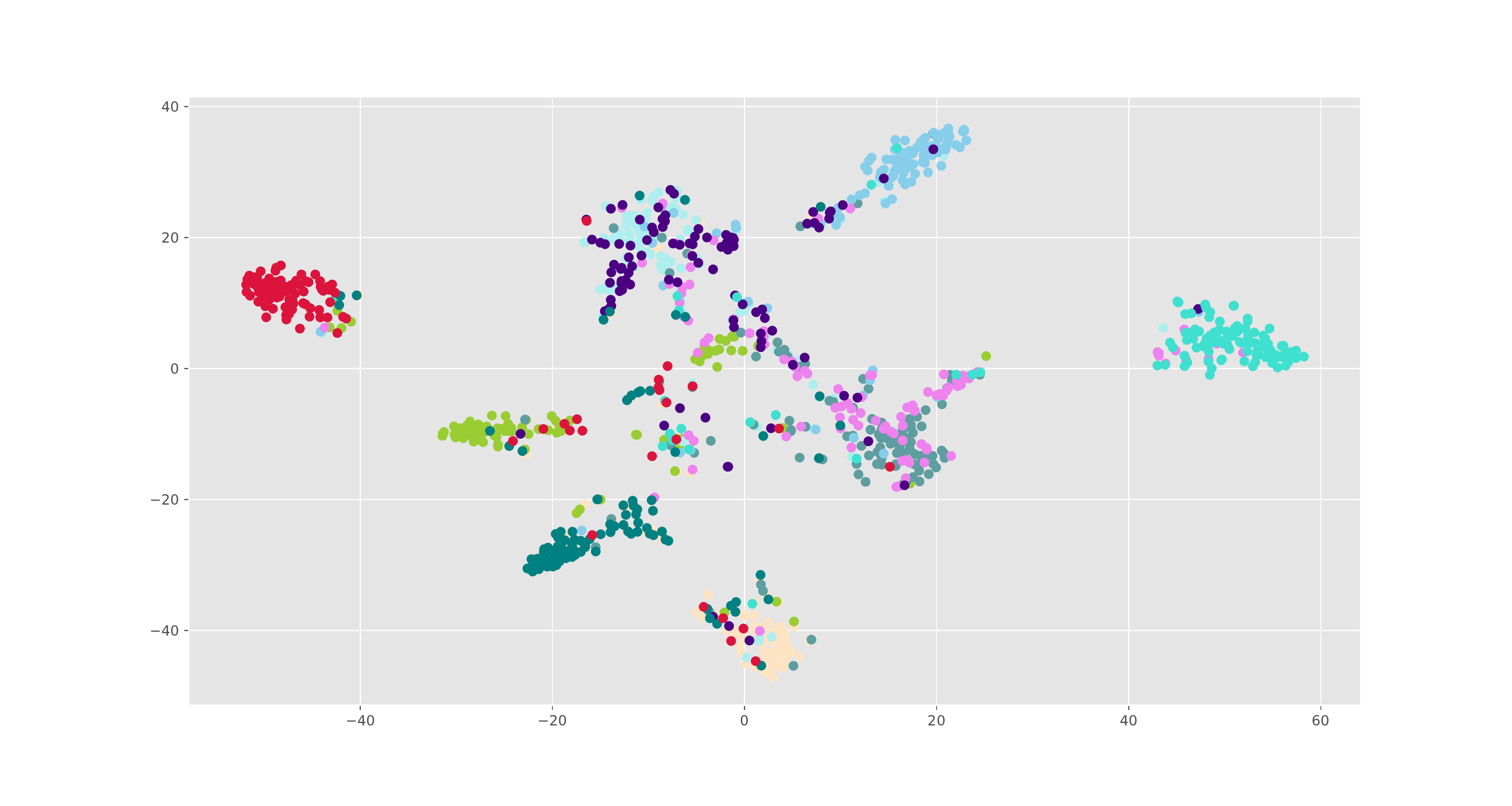},width=6.4cm}}
\end{minipage}
\vfill
\begin{minipage}[b]{.32\linewidth}
\centering
\centerline{\epsfig{figure={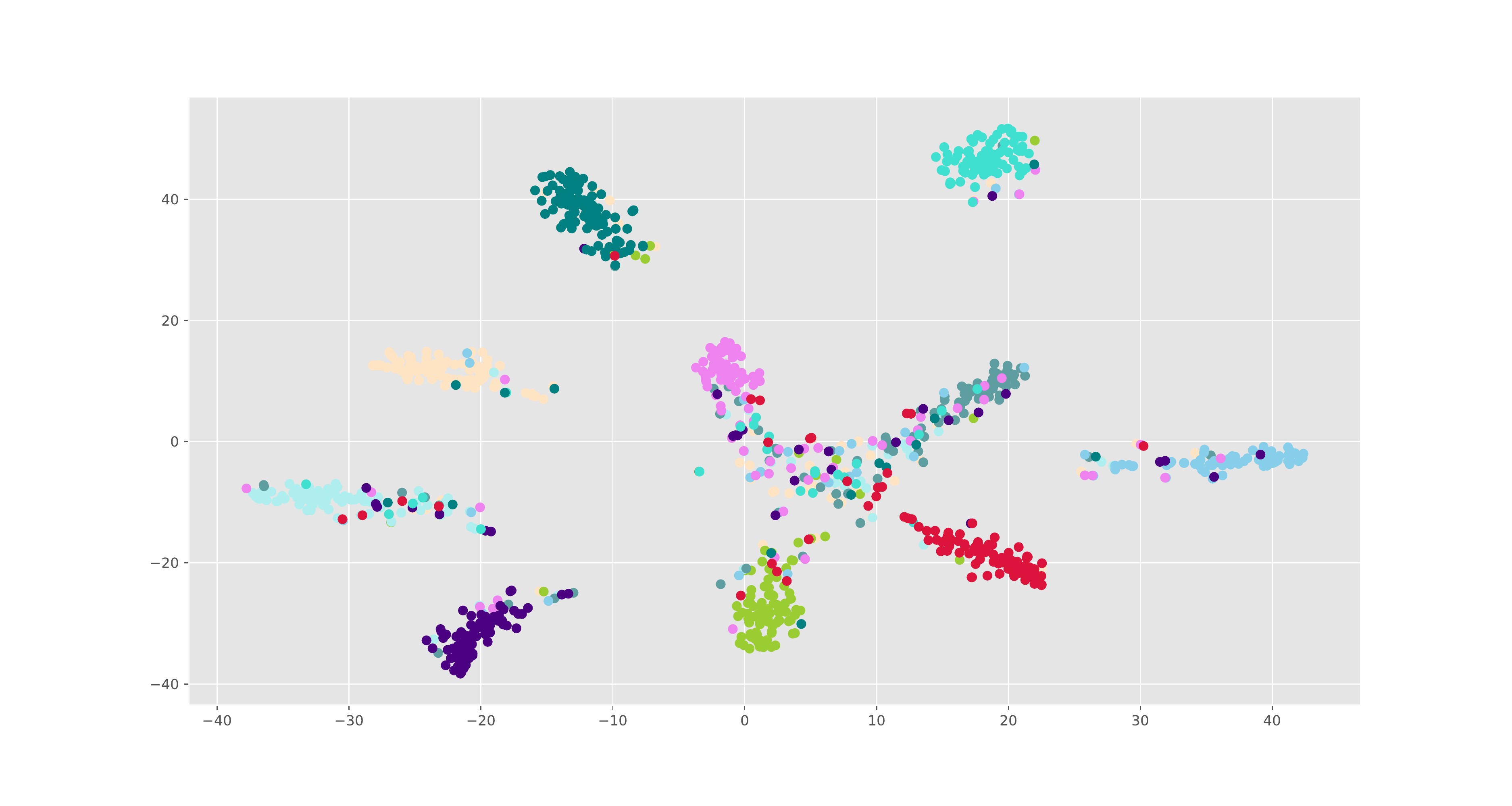},width=6.4cm}}
\end{minipage}
\hfill
\begin{minipage}[b]{.32\linewidth}
\centering
\centerline{\epsfig{figure={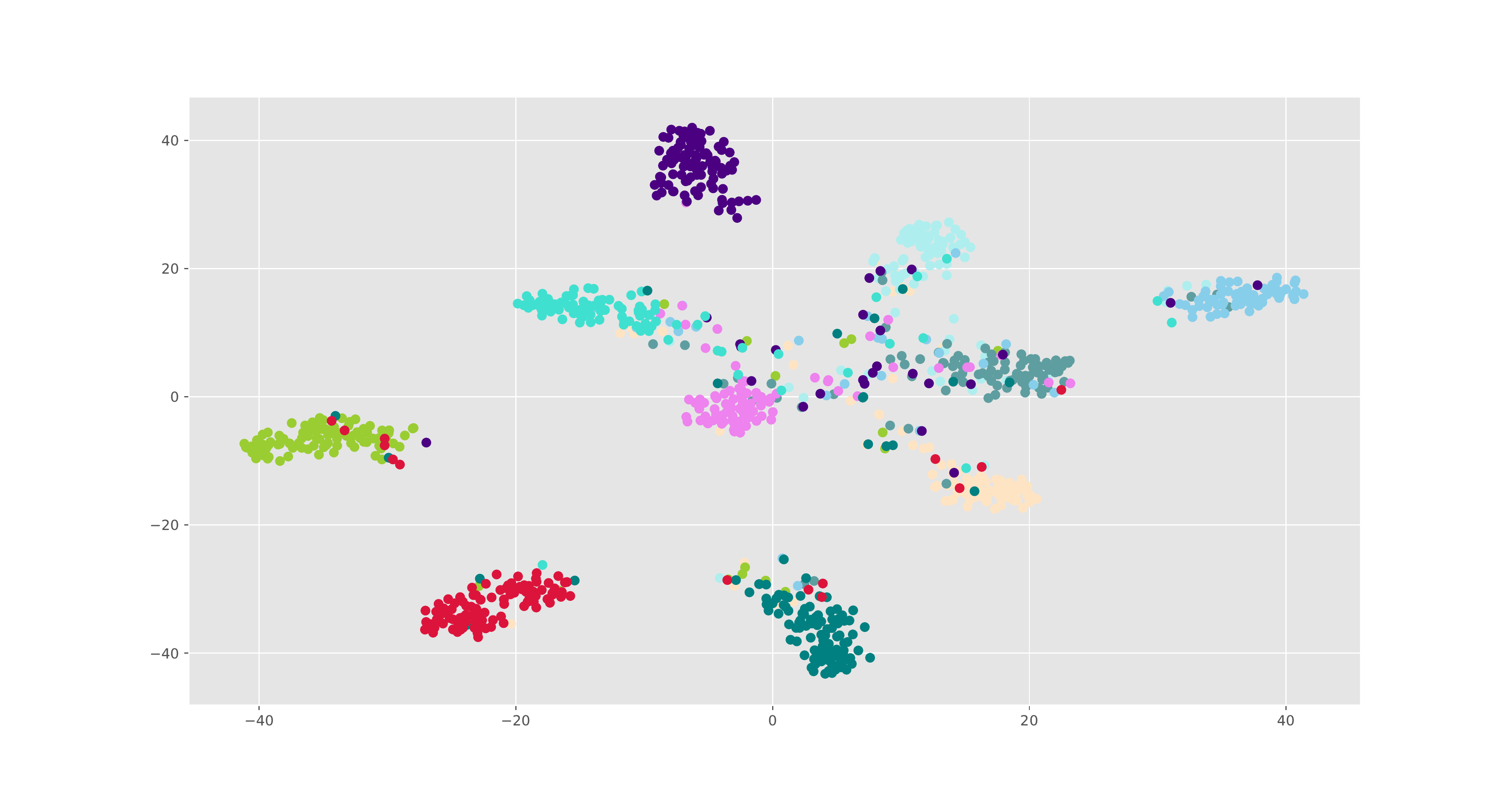},width=6.4cm}}
\end{minipage}
\hfill
\begin{minipage}[b]{.32\linewidth}
\centering
\centerline{\epsfig{figure={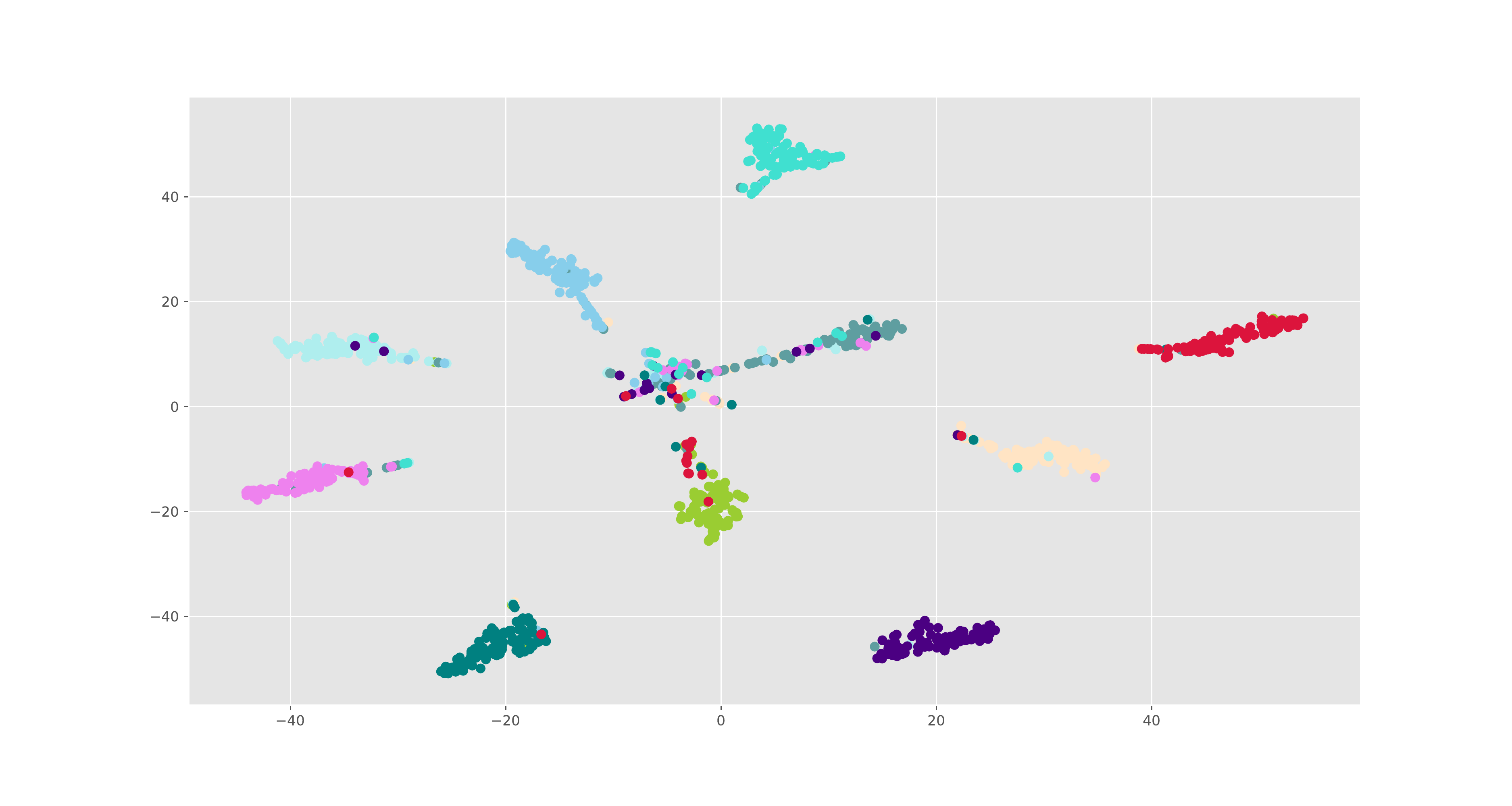},width=6.4cm}}
\end{minipage}
\caption{t-SNE visualisation of penultimate layer Resnet18 features after learning on CIFAR-10 with 40\% symmetric label noise. Top left: model trained by CE. Top middle: model trained by Bootstrap. Top right: model trained by FW. Bottom left: model trained by GCE. Bottom middle: model trained by SCE. Bottom right: model trained by Our learned loss.}
\label{tsne_plot}
\end{figure*}
\subsection{Additional Analysis\label{se:ablation}}
\begin{figure*}[t]
\begin{minipage}[b]{.32\linewidth}
\centering
\centerline{\epsfig{figure={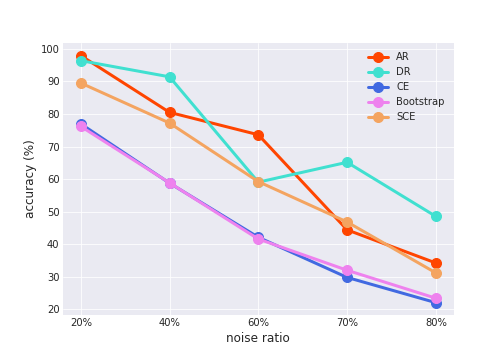},width=6cm}}
\end{minipage}
\hfill
\begin{minipage}[b]{.32\linewidth}
\centering
\centerline{\epsfig{figure={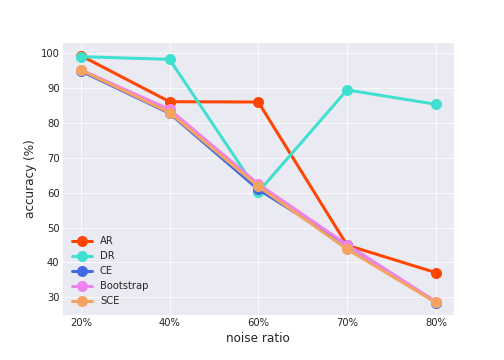},width=6cm}}
\end{minipage}
\hfill
\begin{minipage}[b]{.32\linewidth}
\centering
\centerline{\epsfig{figure={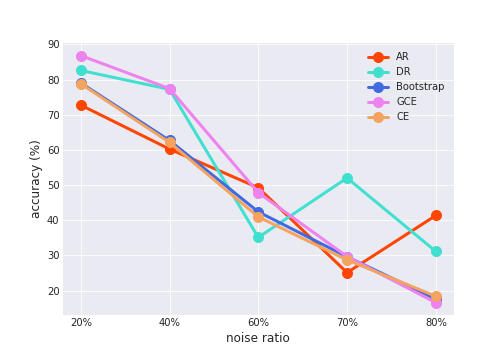},width=6cm}}
\end{minipage}
\cut{\includegraphics[width=.6\textwidth]{taylor_loss/images/2lymnist_save.png}
    \includegraphics[width=.6\textwidth]{taylor_loss/images/4lymnist_save.png}
    \includegraphics[width=.6\textwidth]{taylor_loss/images/vc10_save.png}}
    \caption{Generalisation of learned loss to varying noise-levels. Left: 2MLP-MNIST, Middle: 4CNN-MNIST, Right: VGG11-CIFAR10.}
    \label{fig:noiseStrength}
\end{figure*}

\textbf{Generalisation across noise-levels}\quad We trained our main losses on high levels of label noise ($80\%$-symmetric, $40\%$-asymmetric) as detailed previously, conjecturing that training on a difficult task would be sufficient for generalisation to other tasks with diverse noise conditions, as shown on Clothing1M. To evaluate this more systematically, we apply our $80\%$-symmetric loss on problems with a range of noise levels. From the results in Figure~\ref{fig:noiseStrength} we can see that our loss does tend to provide competitive performance across a range of operating points. 

\textbf{Qualitative analysis of representations}\quad We visualise the feature representation learned by our loss when applied to CIFAR-10 under 40\% symmetric label noise in Figure~\ref{tsne_plot}. We can see that conventional CE applied on noisy labels leads to a very mixed distribution of instances, while our loss leads to quite cleanly separable clusters despite the intense degree of label noise.

\begin{table*}[t]
\centering
\caption{Accuracy (\%) of different robust learners. JoCoR net medium-sized CNN used throughout. Taylor Loss is trained specifically for the target problem. }
\label{tab:mnist}
\resizebox{\textwidth}{!}{
\begin{tabular}{ cccccccccc}
\toprule
& Noise Type & CE (ours) & CE (JoCoR) & GCE & SCE & FW & Bootstrap & JoCoR & Ours \\
\midrule
\parbox[t]{2mm}{\multirow{4}{*}{\rotatebox[origin=c]{90}{MNIST}}} &
 Sym-20\%  & 81.21$\pm$0.53 & 79.56$\pm$0.44  & 97.64$\pm$0.65 & 89.50$\pm$0.44 & 96.85$\pm$0.67 & 76.18$\pm$0.98 & \textbf{98.06$\pm$0.04} & 97.90$\pm$0.12          \\
& Sym-50\%  & 59.51$\pm$0.70 & 52.66$\pm$0.43  & 94.14$\pm$1.32 & 67.38$\pm$0.53 & 94.25$\pm$0.43 & 51.53$\pm$1.56 & 96.64$\pm$0.12          & \textbf{96.71$\pm$0.21} \\
& Sym-80\%  & 22.43$\pm$1.21 & 23.43$\pm$0.31  & 40.57$\pm$0.72 & 31.23$\pm$0.89 & 54.01$\pm$1.82 & 23.46$\pm$0.46 & 84.89$\pm$4.55          & \textbf{89.88$\pm$0.34} \\
& Asy-40\% & 78.73$\pm$1.16 & 79.00$\pm$0.28  & 81.94$\pm$1.22 & 79.87$\pm$0.78 & 90.14$\pm$0.67 & 78.31$\pm$2.34 & 95.24$\pm$0.10          & \textbf{97.38$\pm$0.17} \\
\midrule

\parbox[t]{2mm}{\multirow{4}{*}{\rotatebox[origin=c]{90}{CIFAR-100}}} &
 Symm-20\%  & 39.19$\pm$0.58 & 35.14$\pm$0.44 & 34.66$\pm$0.76 & 35.09$\pm$0.50 & 38.18$\pm$0.76 &  3.53$\pm$0.18 & \textbf{53.01$\pm$0.04} & 51.34$\pm$0.10  \\
& Symm-50\%  & 19.50$\pm$0.43 & 16.97$\pm$0.40 & 10.29$\pm$0.53 & 18.54$\pm$0.29 &  3.25$\pm$0.15 & 18.36$\pm$0.63 & \textbf{43.49$\pm$0.46} & 42.18$\pm$0.27  \\
& Symm-80\%  &  5.56$\pm$0.24 &  4.41$\pm$0.14 &  2.03$\pm$0.36 &  5.75$\pm$0.39 &  6.12$\pm$0.27 &  2.33$\pm$0.13 & 15.49$\pm$0.98 & \textbf{20.20$\pm$0.42}  \\
& Asym-40\% & 30.16$\pm$0.44 & 27.29$\pm$0.25 &  1.32$\pm$0.23 & 27.07$\pm$0.42 &  4.23$\pm$0.51 & 31.72$\pm$0.74 & 32.70$\pm$0.35 & \textbf{36.01$\pm$0.39} \\
\midrule
Avg.Rank & & 5.25 & 6.13 & 5.62 & 5.00 & 4.38 & 6.63 & 1.63 & \textbf{1.38}  \\
\bottomrule

\end{tabular}
}
\end{table*}

\textbf{Dataset-specific loss learning}\quad Our main goal in this paper has been to learn a general purpose robust loss. In this section we examine an alternative use case of applying our framework to train a \emph{dataset-specific} robust loss, in which case better performance could be achieved by customising the loss for the target problem. To achieve this, we now additionally assume a clean subset of data for the target problem is available (unlike the previous experiments, but similarly to several alternative methods in this area~\cite{wei2020combating}) in order to drive loss learning. For this experiment we focus on comparison with JoCoR~\cite{wei2020combating}, since this is the current state-of-the-art model, and in their experiments medium sides networks are applied.  We use the same medium sized CNN architecture as JoCoR for fair comparison, and train our loss to optimize the validation performance. From the results in Table~\ref{tab:mnist}, we can see that our method provides comparable or better performance than state of the art competitor JoCoR. However, this is now at significantly greater cost since the cost of data-specific loss training is not amortizable over multiple tasks as before.

\textbf{Qualitative Analysis and Intuition of Learned Loss}\quad 
To gain some intuition about our loss functions' efficacy, we compare popular standard and robust losses in Figure~\ref{learned_loss_fuc_binary}. Comparing our robust loss, and comparison with the alternatives, we conjecture that there are two properties that impact label-noise robustness in practice: Feedback in response to perceived major prediction errors by the network, and the location of the minima where network predictions maximally satisfy the loss. In the case of a noisy labelled example that the network actually classifies correctly, (e.g., $y_{true}=1$, $y_{label}=0$, $y_{pred}\approx1$), conventional CE aggressively ``corrects'' the network by reporting exponentially large loss. This aggressive feedback can lead to fast training on clean data, but overfitting in noisy data~\cite{zhang2018generalized}. 
Existing robust alternatives MAE~\cite{ghosh2017robust} and GCE~\cite{zhang2018generalized} are explicitly motivated by softening this aggressive ``correction'' compared to CE. Although not explicitly motivated by this, SCE also softens the feedback as shown in the figure. Meanwhile in terms of the minima that best satisfies the loss, conventional CE, as well as SCE, GCE and MAE lead to maximally confident predictions (minima at $0$ or $1$); which, if applied to a noisy label, leads to overfitting. In contrast, label smoothing~\cite{pereyra2017regularizing,wang2019symmetric} improves robustness by inducing softer minima at $[0+\epsilon,1-\epsilon]$ compared to the others' $[0,1]$. However, LS issues the same aggressive correction of large errors as CE, and thus suffers from this accordingly. Only our Taylor loss has learned to exploit both these strategies of less aggressive ``corrections'' and softer targets.

\section{Conclusion}
In this work, we leverage CMA-ES to discover novel loss functions in the defined Taylor series function space. A framework based on domain randomization is developed and instantiated with two variations to enable transferability and generality of the learned loss functions. In order to demonstrate the efficacy of the learned loss functions, we deploy them on to a variety of tasks where the dataset and architecture are different to those seen during the meta-learning process. Comparison with recent work demonstrates the strength of our method empirically. In addition, we show that the proposed method is also able to produce well-behaved models trained with noisy data and these models outperform state-of-the-art models on a range of  tasks. One of the key benefits of our approach is the ability to learn a loss function that can operate on domains with label noise, despite having no clean validation set for that domain---a trait that prominent meta-learning approaches to noise-robust loss function learning do not share~\cite{ren2018learning,shu2019metaWeightNet}. \cut{We will provide an implementation of our method and experiments online.\footnote{Removed for anonymization}}
\clearpage

{\small
\bibliographystyle{ieee_fullname}
\bibliography{6_reference}
}

\end{document}